\documentclass{article}
\PassOptionsToPackage{numbers, compress}{natbib}

\usepackage[preprint]{neurips_2022}

\usepackage[utf8]{inputenc} 
\usepackage[T1]{fontenc}    
\usepackage{hyperref}       
\usepackage{url}            
\usepackage{booktabs}       
\usepackage{amsfonts}       
\usepackage{nicefrac}       
\usepackage{microtype}      
\usepackage{xcolor}         

\usepackage{amsmath}
\usepackage{amsfonts}
\usepackage{mathrsfs}
\DeclareMathOperator*{\argmax}{\arg\!\max}
\usepackage{graphicx}
\usepackage{multirow}
\usepackage{amsmath}
\usepackage{caption}
\usepackage{subfigure}

\usepackage[ruled,linesnumbered]{algorithm2e}

\title{More Interpretable Graph Similarity Computation via Maximum Common Subgraph Inference}

\author{Zixun Lan$^{1}$\thanks{Equal contribution} \qquad   Binjie Hong$^{2\,*}$ \qquad   Ye Ma$^{3}$\qquad   Fei Ma$^1$\thanks{corresponding author}\\
\\              
$^1$\ Department of Applied Mathematics, School of Science\\
$^2$\ Department of Information and Computing Science, School of Advanced Technology\\
$^3$\ Department of Financial and Actuarial Mathematics, School of Science\\
Xi’an Jiaotong-Liverpool University,
SIP, 215123 Suzhou, China\\
\\
\{zixun.lan19, binjie.hong19\}@student.xjtlu.edu.cn, \{ye.ma, fei.ma\}@xjtlu.edu.cn}

\begin{document}

\maketitle

\begin{abstract}
Graph similarity measurement, which computes the distance/similarity between two graphs, arises in various graph-related tasks. Recent learning-based methods lack interpretability, as they directly transform interaction information between two graphs into one hidden vector and then map it to similarity. To cope with this problem, this study proposes a more interpretable end-to-end paradigm for graph similarity learning, named Similarity Computation via Maximum Common Subgraph Inference (INFMCS). Our critical insight into INFMCS is the strong correlation between similarity score and Maximum Common Subgraph (MCS). We implicitly infer MCS to obtain the normalized MCS size, with the supervision information being only the similarity score during training. To capture more global information, we also stack some vanilla transformer encoder layers with graph convolution layers and propose a novel permutation-invariant node Positional Encoding. The entire model is quite simple yet effective. Comprehensive experiments demonstrate that INFMCS consistently outperforms state-of-the-art baselines for graph-graph classification and regression tasks. Ablation experiments verify the effectiveness of the proposed computation paradigm and other components. Also, visualization and statistics of results reveal the interpretability of INFMCS.
\end{abstract}

\section{Introduction}

Graph similarity measurement, which is to compute distance/similarity between two graphs, is a fundamental problem in graph-related tasks. It arises in a variety of real-world applications, such as graph search in graph-based database \cite{yan2002gspan}, malware detection \cite{wang2019heterogeneous}, brain data analysis \cite{ma2019deep}, etc. Graph Edit Distance (GED) \cite{bunke1983distance} and Maximum Common Subgraph (MCS) \cite{bunke1998graph} are two domain-agnostic graph similarity metrics, yet exact computation of both are known to be NP-hard \cite{zeng2009comparing}. For instance, no algorithm can compute the exact GED between graphs of more than 16 nodes within a reasonable time so far \cite{blumenthal2020exact}. This problem has motivated interest in approximation algorithms, with a recent surge in graph similarity learning methods \cite{li2019graph,bai2019simgnn,bai2020learning,ling2021multilevel,zhang2021h2mn,xu2021graph,lan2021sub,lan2022aednet}.

Recent methods improve performance by capturing node-level or subgraph-level interactions. The initial way is to encode each graph as a graph-level fixed-length vector via Graph Neural Networks (GNNs) and then combine the two vectors of both input graphs to predict similarity. However, the actual difference between two graphs often arises from very small local substructures, resulting in the graph-level fixed-length vector being difficult to contain local information \cite{bai2019simgnn}. To alleviate this problem, GMN \cite{ li2019graph}, MGMN \cite{ling2021multilevel} and H2MN \cite{zhang2021h2mn} derive node-level and graph-level embeddings containing interaction information of different scales through the cross-graph attention (propagation), and then convert these embeddings to one hidden vector (e.g. concatenation between two graph-level embeddings). SimGNN \cite{bai2019simgnn} and GraphSim \cite{bai2020learning} derive the corresponding hidden vector by applying the convolution operation to the pairwise node similarity matrix or extracting its histogram features respectively. Ultimately, all models map the hidden vector to the ground-truth similarity.

Previous methods lack interpretability despite the exploitation of interaction information. It is unclear what the final hidden vector represents and how to map it to ground truth. In natural language processing \cite{vashishth2019attention}, it has been demonstrated that models with more interpretability tend to boost performance. Many tasks use attention, such as machine translation \cite{luong2015effective}, language modelling \cite{liu2018learning}, abstractive summarization \cite{ma2021global}. Attention not only provides interpretability \cite{wang2016attention,lin2017structured,ghaeini2018interpreting}, but also benefits the performance of models . Without loss of generality, the graph similarity learning model with a more reasonable and interpretable paradigm can generally capture more critical information and filter out interference information, thus outperforming the less interpretable one.

\begin{figure}
	\centering
	\includegraphics[width=1\linewidth]{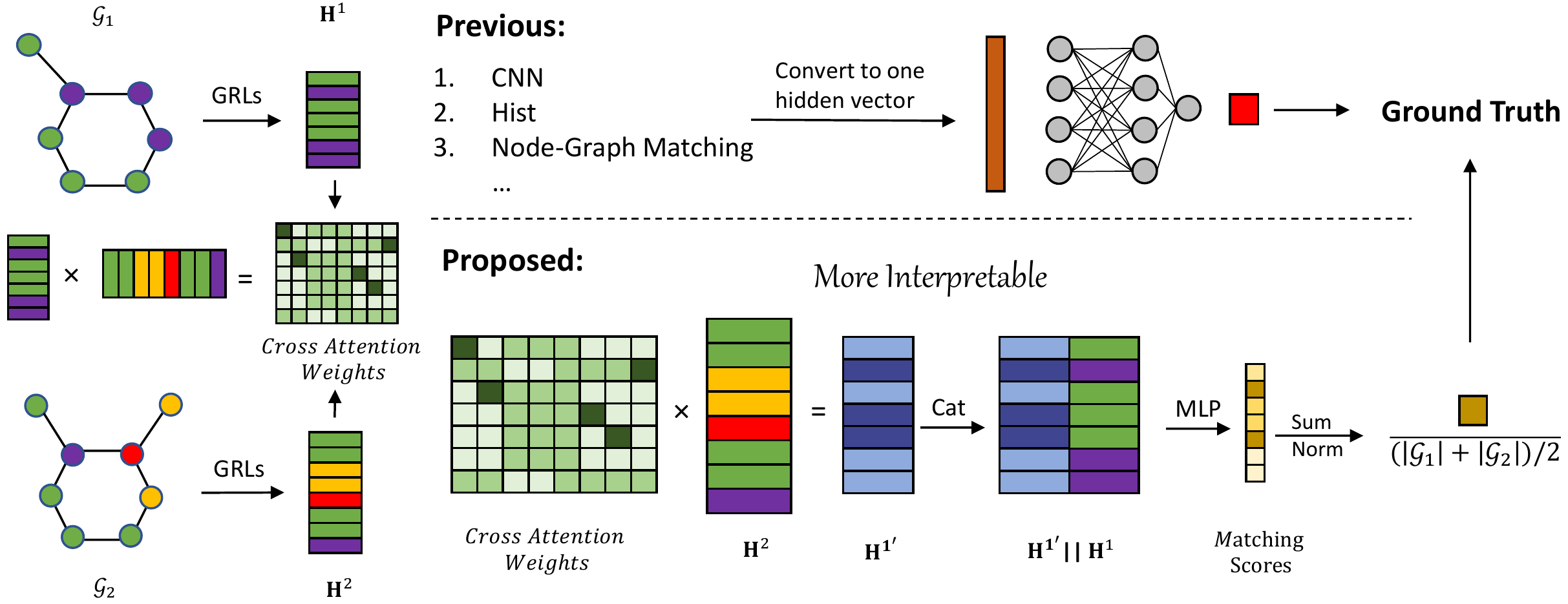}	\caption{The entire flow of the proposed INFMCS framework. ({\bf Previous:}) The previous method extracts hidden vectors from interactions of different scales and then maps them to the ground-truth similarity. It is less interpretable, and mapping hidden vectors to ground truth is also agnostic. ({\bf Proposed:}) Our proposed model follows that the more similar a pair of graphs is, the greater the ratio of the MCS size to the pair's average size is. Although only the ground-truth similarity score is used during training, we can infer the MCS from inside the model.}
	\label{f1}
\end{figure}

To cope with this limitation, this study proposes a more interpretable end-to-end paradigm for graph similarity learning, named Similarity Computation via Maximum Common Subgraph Inference (INFMCS). Commonly, the more significant proportion of MCS to the average size of graph pairs \cite{bai2020learning}, i.e. normalized MCS size ($\mathrm{nMCS}$)\footnote{$\mathrm{nMCS}\left(\mathcal{G}_{1}, \mathcal{G}_{2}\right)= \frac{\left|\mathrm{MCS}\left(\mathcal{G}_{1}, \mathcal{G}_{2}\right)\right|}{\left(\left|\mathcal{G}_{1}\right|+\left|\mathcal{G}_{2}\right|\right) / 2}$. In this paper, we always view the graph with the smaller size as $\mathcal{G}_{1}$, because the MCS size between two graphs is less than or equal to the size of the smaller graph.}, the more similar the two graphs are. According to this fact, we infer Maximum Common Subgraph (MCS) implicitly and then obtain the normalized MCS size in an end-to-end fashion. First, we perform message passing from $\mathcal{G}_{2}$ to $\mathcal{G}_{1}$ by the modified cross-graph attention mechanism, thereby obtaining $|\mathcal{G}_{1}|$ pairs of node-level embeddings. In each pair, one embedding represents one node in $\mathcal{G}_{1}$ and the another embedding represents one node in $\mathcal{G}_{2}$ most likely matching the former. After concatenating the two embeddings in the embedding pair, we use MLP to transform the concatenation to the matching score between zero and one, where one and zero represent that two nodes are matched and not matched respectively. Finally, we add up the $|\mathcal{G}_{1}|$ matching scores as the predicted MCS size and then normalize it to derive the predicted similarity score. The entire process is optimized by either GED\footnote{$\mathrm{nGED}\left(\mathcal{G}_{1}, \mathcal{G}_{2}\right)=\exp(\frac{-\mathrm{GED}\left(\mathcal{G}_{1}, \mathcal{G}_{2}\right)}{\left(\left|\mathcal{G}_{1}\right|+\left|\mathcal{G}_{2}\right|\right) / 2}).$}/MCS normalized similarity or graph-graph classification label end-to-end. 
We also stack a few vanilla transformer encoder layers \cite{vaswani2017attention} with graph convolution layers to capture more global information, called Graph Convolution with Transformer (GCwT). Unlike the inherent order of sentences in natural language \cite{vaswani2017attention}, graphs are permutation-invariant, resulting in no order for nodes. Thus we propose a novel Positional Encoding based on permutation-invariant node ordering. Comprehensive experiments demonstrate that INFMCS consistently outperforms state-of-the-art baselines for graph-graph classification and regression tasks. Ablation experiments verify the effectiveness of individual components, including the proposed graph similarity learning paradigm and GCwT with novel Positional Encoding. In brief, we highlight our main contributions as follows:

\begin{itemize}
    \item We propose a more interpretable end-to-end paradigm for graph similarity learning. The interpretability is derived from inferring MCS implicitly.
    
    \item To our best knowledge, the proposed Positional Encoding based on permutation-invariant node ordering is the first being proposed and used in the graph-transformer related model for graph similarity task. Like the order of tokens in sentences, permutation-invariant node ordering is also an essential graph feature.
    
    \item We perform comprehensive experiments on six graph-graph regression datasets and two graph-graph classification datasets to verify the effectiveness of the INFMCS. Ablation experiments verify the effectiveness of individual components. Also, the case study and visualization demonstrate interpretability.
\end{itemize}

\section{Related Work}

Initial methods, such as SMPNN \cite{riba2018learning}, GCNMEAN and GCNMAX \cite{ktena2017distance} directly encode each graph as a graph-level fixed-length vector via GNNs and then only use graph-level interaction to predict similarity. After that, more models were proposed to exploit node-level or subgraph-level interactions by degrees. GMN \cite{li2019graph} uses cross-graph attention to derive node-level embeddings that contain another graph's information. SimGNN \cite{bai2019simgnn} and GraphSim \cite{bai2020learning} derive the corresponding hidden vector differently by applying the convolution operation to the pairwise node similarity matrix or extracting its histogram features. \cite{xu2021graph} first partitions graphs and then conducts node-wise comparison among subgraphs. MGMN \cite{ling2021multilevel} designs node-graph matching layers by comparing each node's representation of one graph with the other whole graph representation. After converting graph to hypergraph via random walk or K-hop neighbourhood, H2MN \cite{zhang2021h2mn} utilizes hypergraph convolution and subgraph matching blocks to predict similarity.

Compared with previous methods, our proposed INFMCS is simpler. INFMCS does not need to compute node-wise interactions per layer \cite{li2019graph,bai2019simgnn} but only uses the last layer of node-level embeddings to capture cross-graph interactions. Second, ours does not need to consider multi-scale matching scores of $|\mathcal{G}_{1}| \times |\mathcal{G}_{2}|$ pairs \cite{ling2021multilevel}, only calculates $|\mathcal{G}_{1}|$ matching scores. Third, ours does not need to preprocess graph data, such as graph partition \cite{xu2021graph} and hypergraph construction \cite{zhang2021h2mn}. Although our method avoids the above computational burden, it achieves good performance.

\section{Model Design}

Our approach follows the hypothesis that the more similar a pair of graphs is, the greater the ratio of MCS size between graphs to the pair's average size is. To achieve this, we first derive $|\mathcal{G}_{1}|$ pairs of node embeddings via cross-graph attention and then transform them into matching scores. Ideally, the sum of $|\mathcal{G}_{1}|$ matching scores is precisely equal to the size of MCS. Finally, we normalize the sum of these scores to predict the similarity. We also proposed a Graph Convolution with Transformer based on permutation-invariant Positional Encoding to fill a gap between the shallow GCN and the sizeable receptive field. The overall process is end-to-end and is outlined in Figure 1. Before describing the modules, we introduce relevant preliminaries that set the background for the remainder of the paper.

\textbf{Graph Similarity Learning} Given a pair of input graphs $(\mathcal{G}_{1}, \mathcal{G}_{2})$, the aim of graph similarity learning is to produce a similarity score $y=s\left(\mathcal{G}_{1}, \mathcal{G}_{2}\right) \in \mathcal{Y}$. The graph $\mathcal{G}_{1}=\left(\mathcal{V}_{1}, \mathcal{E}_{1}\right)$ is represented as a set of $N$ nodes $v_{i} \in \mathcal{V}_{1}$ with a feature matrix $X_{1} \in \mathcal{R}^{N \times d}$, edges $\left(v_{i}, v_{i^{r}}\right) \in \mathcal{E}_{1}$ formulating an adjacency matrix $A_{1} \in \mathcal{R}^{N \times N}$. Similarly,
the second graph $\mathcal{G}_{2}=\left(\mathcal{V}_{2}, \mathcal{E}_{2}\right)$ can be represented in the same way. For graph-graph classifcation task, the scalar $y$ represents the class label, i.e., $y \in \mathcal{Y}=\{0,1\}$; for graph-graph regression task, the scalar $y$ measures the graph similarity, i.e., $y \in \mathcal{Y}=\left[0,1\right]$.

\subsection{Similarity Computation}

Notably, we always view the graph with the smaller size as $\mathcal{G}_{1}$, since the MCS size between two graphs is less than or equal to the size of the smaller graph in this paper. Given the node representations of the last layer of the graph representation learning $\mathbf{H}^{1}=[\mathbf{h}_{1}^{1};\mathbf{h}_{2}^{1}; \cdots;\mathbf{ h}_{|\mathcal{V}_{1}|}^{1}] \in \mathcal{R}^{|\mathcal{V}_{1}| \times d}$ for $\mathcal{G}_{1}$ and $\mathbf{H }^{2}=[\mathbf{h}_{1}^{2};\mathbf{h}_{2}^{2}; \cdots;\mathbf{h}_{|\mathcal {V}_{2}|}^{2}] \in \mathcal{R}^{|\mathcal{V}_{2}| \times d}$ for $\mathcal{G}_{2}$, we pass the message from $\mathcal{G}_{2}$ to $\mathcal{G}_{1}$ by modified cross-graph attention $a_{ij}$, and then obtain the representation $\mathbf{h}_{i^{\prime}}^{1}\in \mathcal{R}^{1 \times d}$ of node $v_{j} \in \mathcal{V}_{2}$ that most likely matches node $v_{i} \in \mathcal{V}_{1}$:
\begin{equation}
\label{e4}
\begin{aligned}
a_{ij} =\frac{\exp \left(s_{h}\left(\mathbf{h}_{i}^{1}, \mathbf{h}_{j}^{1}\right) \times \tau_{*}^{-1}\right)}{\sum_{j^{\prime}} \exp \left(s_{h}\left(\mathbf{h}_{i}^{1}, \mathbf{h}_{j^{\prime}}^{1}\right)\times \tau_{*}^{-1}\right)}, 
\mathbf{h}_{i^{\prime}}^{1} &=\sum_{j}a_{ij}\mathbf{h}_{i}^{(t)}, 
\end{aligned}
\end{equation}
where $s_{h}$ is  a vector space similarity metric, like Euclidean or cosine similarity. In order to discretize $a_{ij}$, we add a learnable parameter $\tau_{*} \in (0,1]$. In other words, it makes the weight $a_{ij}$ of one node $v_{j} \in \mathcal{V}_{2}$ ($j=\argmax_{j^{\prime}}a_{ij^{\prime}}$) tend to one and the others tend to zero due to $\sum_{j^{\prime}}a_{ij^{\prime}}=1$. Thus, $\mathbf{h}_{i^{\prime}}^{1}$ represents the representation of node corresponding to node $v_{i}$ with the highest probability.

After concatenating $\mathbf{h}_{i}^{1}$ with $\mathbf{h}_{i^{\prime}}^{1}$, we transform the concatenation to the matching score $s_{i}$ by MLP. Ideally, the sum of $|\mathcal{G}_{1}|$ matching scores is precisely equal to the size of MCS. Finally, we normalize the sum of these predicted scores to compute similarity $\hat{y}_{i}$:
\begin{equation}
\label{e5}
\hat{y}=\frac{\sum_{i}s_{i}}{(|\mathcal{G}_{1}|+|\mathcal{G}_{2}|) / 2}, s_{i} = \mathrm{sigmoid}\left(\mathrm{MLP}\left(\mathbf{h}_{i}^{1} \| \mathbf{h}_{i^{\prime}}^{1} \right)\right).
\end{equation}

The loss functions are defined as follows:
\begin{equation}
\label{e6}
\begin{aligned}
\mathcal{L}_{c}=-\frac{1}{|\mathrm{D}|} \sum_{i=1}^{|\mathrm{D}|} y_{i} \log \left(\hat{y}_{i}\right)+\left(1-y_{i}\right) \log \left(1-\hat{y}_{i}\right) \, or \, \mathcal{L}_{r}=\frac{1}{|\mathrm{D}|} \sum_{i=1}^{|\mathrm{D}|}\left(y_{i}-\hat{y}_{i}\right)^{2},
\end{aligned}
\end{equation}
where $\mathcal{L}_{c}$ represents the binary cross-entropy loss for the graph-graph classification task and $\mathcal{L}_{r}$ is the mean square error loss for the graph-graph regression task. $y_{i}$ denotes the ground-truth supervision information, and $|\mathrm{D}|$ is the size of the dataset.

\subsection{Graph Convolution with Transformer}

Before exploiting node-wise interactions, we need to obtain node-level embeddings as in the previous method. GCN \cite{kipf2016semi} is the most popular spatial graph convolution. In this study, we use it to compute node-level embeddings. For simplicity, we denote the encoding process by $\mathrm{GCN}(\cdot)$ and describe architectural details in the appendix. The $\mathrm{GCN}$ computes node representations $\mathbb{H} \in \mathcal{R}^{|\mathcal{V}| \times d}$ via
\begin{equation}
\label{e1}
\begin{aligned}
\mathbb{H}=\left[\mathrm{h}_{1};\mathrm{h}_{2}; \cdots;\mathrm{h}_{|\mathcal{V}|}\right], \mathrm{h}_{i}=\mathrm{GCN}\left(\mathcal{G},\mathbf{x}_{i},\left\{\mathbf{x}_{i j}\right\}_{j \in \mathcal{N}(i)}\right),
\end{aligned}
\end{equation}
where $\mathrm{h}_{i} \in \mathcal{R}^{1 \times d}$ and $\mathcal{N}(i)$ denotes the node representation and neighbors of node $v_{i}$ respectively. Graph similarity learning requires not only node embeddings to perceive local information but also node representations to contain global information due to many subtle differences across the whole graph \cite{bai2020learning,zhang2021h2mn}. However, over-smoothing \cite{rong2019dropedge,kipf2016semi} constrains graph convolution from stacking multiple layers, resulting in a gap between the shallow GCN and the sizeable receptive field.

To fill this gap, we stack some vanilla transformer encoder layers \cite{vaswani2017attention} with graph convolution layers to capture more global information naturally. The global perception ability of Transformer comes from the self-attention mechanism, which internally calculates the correlation between embeddings. The attention weight matrix is equivalent to constructing a fully connected graph, and then message passing is based on this fully connected structure. For sentence representation \cite{vaswani2017attention}, extensive experiments show the importance of Positional Encoding. Sentences with the same tokens but in different order have different semantics (Figure 2), which shows that order is also an inherent feature of sentences. However, graphs are permutation-invariant, resulting in no order for nodes. Thus, we propose a permutation-invariant node ordering $\mathbf{C} \in \mathcal{R}^{|\mathcal{V}|}$ based on closeness centrality \cite{wasserman1994social}:
\begin{equation}
\label{e2}
\mathbf{C}=\mathrm{argtop}|\mathcal{V}|\left(\left[\mathrm{c}_{1},\mathrm{c}_{2}, \cdots, \mathrm{c}_{|\mathcal{V}|}\right]\right), \mathrm{c}_{i}=\frac{n-1}{|\mathcal{V}|-1} \frac{n-1}{\sum_{j=1}^{n-1} d(j, i)},
\end{equation}
where $\mathrm{c}_{i}$ is the closeness centrality of node $v_{i}$. It is the reciprocal of the average shortest path distance to $v_{i}$ over all other nodes, and higher closeness values indicate higher centrality. Besides, $n$ is the number of nodes in the connected part of the graph containing the node $v_{i}$ and $\frac{n-1}{|\mathcal{V}|-1}$ is the proportion of this connected component in the entire graph. Hence, Eq. \ref{e2} can be generalized to graphs with more than one connected component, where the size of the connected component scales each node. $\mathrm{argtop}|\mathcal{V}|(\cdot)$ calculates the rank of each element in the vector in descending order. For example, $\mathrm{argtop}|\mathcal{V}|(\left[0.4,0.6,0.1,0.9\right])=[2,1,3,0]$. Our model also generalizes to edge representations. We convert the original graph to a line graph and then compute the permutation-invariant node order since the edges in the original graph construct the nodes in the line graph.

\begin{figure}
	\centering
	\includegraphics[width=1\linewidth]{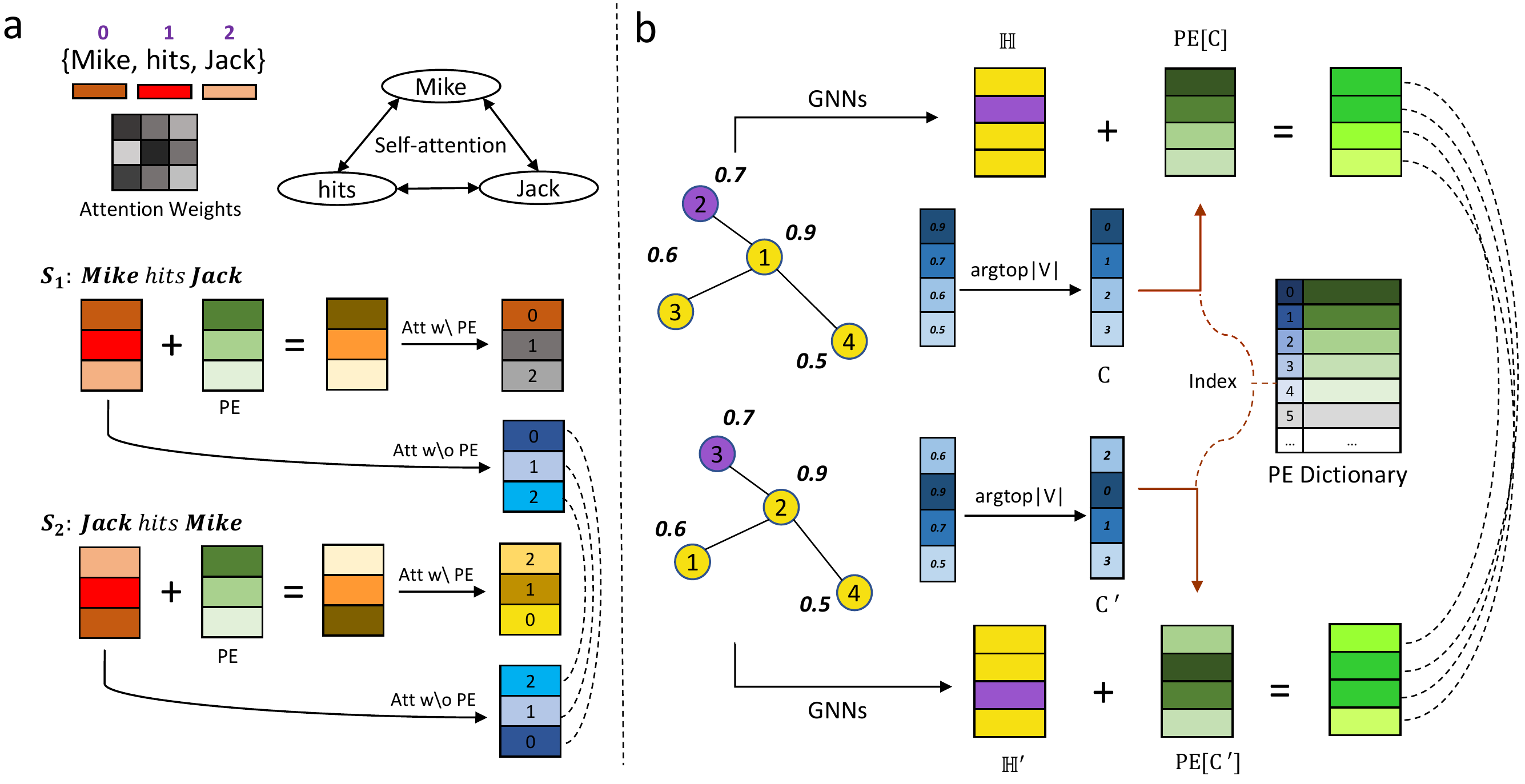}
	\caption{{\bf a:} Sentences with the same tokens but in different order have different semantics, which shows that order is also an inherent feature of sentences. For example, sentence $s_{1}$ and sentence $s_{2}$ have different semantics, while the same token gets the identical representation by directly using self-attention without PE. {\bf b:} Our proposed Positional Encoding is permutation-invariant. Even if the nodes in the graph are re-permutated, the Positional index for these nodes does not change.}
	\label{f2}
\end{figure}

We denote the vanilla transformer encoder by $\mathrm{TranformerEncoder}(\cdot)$ for simplicity and describe architectural details in appendix. Given a learnable Positional Encoding dictionary $\mathbf{PE} \in \mathcal{R}^{m \times d}(m \gg |\mathcal{V}|)$, final node representations $\mathbf{H} \in \mathcal{R}^{|\mathcal{V}| \times d}$ is derived by Eq. \ref{e3}:
\begin{equation}
\label{e3}
\begin{aligned}
\mathbf{H}=\mathrm{TranformerEncoder}(\mathscr{H}), 
\mathscr{H} = \mathbb{H} + \mathbf{PE}\left[\mathbf{C}\right],
\end{aligned}
\end{equation}
where $\mathbf{PE}\left[\mathbf{C}\right] \in \mathcal{R}^{|\mathcal{V}| \times d}$ is Positional Encoding according to the node ordering $\mathbf{C}$.

\section{Evaluation}
In this section, we systematically evaluate the performance of our INFMCS with comparison to recently proposed state-of-the-art approaches for both the graph-graph classification and graph-graph regression tasks, and with significant goals of addressing the following questions: {\bf Q1: }How effective, efficient and robust is INFMCS compared to the state-of-the-art approaches under MCS/GED metric? {\bf Q2: }How does the proposed similarity computation paradigm and GCwT with permutation-invariant Positional Encoding improve performance? {\bf Q3: }Does INFMCS have stronger interpretability?

{\bf Data }For graph-graph classification task, we use {\bf FFmpeg}\footnote{https://ffmpeg.org/} and {\bf OpenSSL}\footnote{https://www.openssl.org/} \cite{ling2021multilevel} as datasets, where each graph denotes binary function's control flow graph (CFG). Therefore, we take two CFGs compiled from the same source code as positive samples, i.e., $s\left(\mathcal{G}_{1}, \mathcal{G}_{2}\right)=1$, and the dissimilar CFGs compiled from different source code, i.e., $s\left(\mathcal{G}_{1}, \mathcal{G}_{2}\right)=0$. Moreover, we split each dataset into 3 sub-datasets according to the graph size in order to investigate the impact of graph size. For graph-graph regression task, we employ three real datasets and three sythetic datasets, including {\bf AIDS(2-15)}, {\bf LINUX(2-15)}, {\bf PTC\_MM(all)}, {\bf BA100}, {\bf BA200} and {\bf BA300}. We extract graphs from the original datasets ({\bf AIDS \cite{riesen2008iam}}, {\bf LINUX \cite{wang2012efficient}}, {\bf PTC\_MM \cite{helma2001predictive}})\footnote{https://chrsmrrs.github.io/datasets/docs/datasets/} and construct the above three real datasets, where the values in parentheses indicate the size range of extracted graph. To verify the performance on large graphs, we also use the Barabási–Albert model \cite{jeong2003measuring} to generate sythetic graphs. We generated three sythetic datasets with graph sizes around 100, 200, and 300 respectively, called {\bf BA100}, {\bf BA200} and {\bf BA300}. Detailed descriptions and statistics of both real and synthetic datasets can be found in the appendix.

{\bf Evaluation }For the graph-graph classification task, we use \textit{Area Under the Curve (AUC)} \cite{bradley1997use} to evaluate the model. For the graph-graph regression task, we use \textit{averaged Mean Squared Error (mse), Spearman’s Rank Correlation Coefficient ($\rho$)} \cite{spearman1961proof} and \textit{Precision at k (p@k)} to test the accuracy and ranking performance. To avoid information leakage, we let the model compute the similarity between the query graph and each graph in the original test set to rank.

{\bf Baselines }We use eight SOTA learning-based methods as baselines, including GMN \cite{li2019graph}, SimGNN \cite{bai2019simgnn}, GraphSim \cite{bai2020learning}, MGMN \cite{ling2021multilevel}, H2MN \cite{zhang2021h2mn}, GOTSim \cite{doan2021interpretable}, PSimGNN \cite{xu2021graph}, EMBAVG \cite{bai2020learning} and SMPNN \cite{riba2018learning}. Besides methods PSimGNN \cite{xu2021graph} and EMBAVG \cite{bai2020learning}, we use the source code released by the authors for the regression task. We reproduced the above two baselines according to the original papers. We also tune their hyperparameters on the validation set on the regression task. For the classification task, we use the AUC scores reported in \cite{zhang2021h2mn} as the results for each baseline. We also use five classical algorithms as baselines to compare running time, including A* \cite{riesen2013novel}, MCSPLIT \cite{mccreesh2017partitioning}, BEAM \cite{neuhaus2006fast}, HUNGARIAN \cite{riesen2009approximate}, VJ \cite{fankhauser2011speeding} and HED \cite{fischer2015approximation}.

{\bf Implementation Settings }Our proposed INFMCS is implemented with Deep Graph Library (DGL) \cite{wang2019dgl} and Pytorch \cite{paszke2019pytorch}. Regarding the model's hyperparameters, we fix the number of GCN layers to 3. We search the number of Transformer layers $L \in \{2, 4, 6, 8 \}$ and the dimension of hidden layers $d \in \{128, 256, 512 \}$, where the dimension of all hidden layers is set to the same. More details about the hyper-parameter search can be found in the appendix. We conduct all the experiments on a machine with an Intel Xeon 4114 CPU and two Nvidia Titan GPU. As for training, we use the Adam algorithm for optimization \cite{kingma2014adam} and fix the initial learning rate to 0.001. The proposed model is trained on real datasets for 100 epochs with a batch size of 128 and on synthetic datasets for 30 epochs with a batch size of 32. Checkpoints are saved for each epoch to select the best checkpoints on the evaluation set. The source code can be found in the supplementary materials.

\begin{table}[]
\caption{Graph-Graph classification results (AUC score) with standard deviation (in percentage).}
\label{t_cls}
\centering
\begin{tabular}{c|lll|lll}
\hline
\multirow{2}{*}{Datasets} & \multicolumn{3}{c|}{FFmpeg}                                                                               & \multicolumn{3}{c}{OpenSSL}                                                                              \\ \cline{2-7} 
                        & \multicolumn{1}{c}{{[}3, 200{]}} & \multicolumn{1}{c}{{[}20, 200{]}} & \multicolumn{1}{c|}{{[}50, 200{]}} & \multicolumn{1}{c}{{[}3, 200{]}} & \multicolumn{1}{c}{{[}20, 200{]}} & \multicolumn{1}{c}{{[}50, 200{]}} \\ \hline
\multicolumn{1}{l|}{SimGNN}                                        & 95.38±0.76                       & 94.32±1.01                        & 93.45±0.54                         & 95.96±0.31                       & 93.38±0.82                        & 94.25±0.85                        \\
\multicolumn{1}{l|}{GMN}                                             & 94.15±0.62                       & 95.92±1.38                        & 94.76±0.45                         & 96.43±0.61                       & 93.03±3.81                        & 93.91±1.65                        \\
\multicolumn{1}{l|}{GraphSim}                                        & 97.46±0.30                       & 96.49±0.28                        & 94.48±0.73                         & 96.84±0.54                       & 94.97±0.98                        & 93.66±1.84                        \\
\multicolumn{1}{l|}{MGMN}                                            & 98.07±0.06                       & 98.29±0.10                        & 97.83±0.11                         & 96.90±0.10                       & 97.31±1.07                        & 95.87±0.88                        \\
\multicolumn{1}{l|}{PSimGNN}                                         & 96.67±0.54                       & 96.86±0.95                        & 95.23±0.15                         & 96.10±0.46                       & 94.67±1.30                        & 93.46±1.59                        \\
\multicolumn{1}{l|}{GOTSim}                                         & 96.93±0.34                       & 97.01±0.52                        & 95.65±0.31                         & 97.87±0.49                       & 96.42±1.89                        & 95.97±1.06                        \\
\multicolumn{1}{l|}{H2MN}                                            & 98.28±0.20                       & 98.54±0.14                        & 98.30±0.29                         & 98.27±0.16                       & 98.47±0.38                        & 97.78±0.75                        \\ \hline
\multicolumn{1}{l|}{INFMCS}                                         & \textbf{98.49±0.09}              & \textbf{99.36±0.13}               & \textbf{99.48±0.20}                & \textbf{98.34±0.20}                    & \textbf{99.14±0.31}               & \textbf{99.26±0.45}               \\ \hline
\end{tabular}
\end{table}

\subsection{Overall Performance}

{\bf Graph-Graph Classification Task }We operate the training process five times and report the mean and standard deviation in \textit{AUC}. Our method is straightforward and achieves state-of-the-art performance on both datasets under all settings. The graph-graph classification performance is illustrated in Table \ref{t_cls}. We have two observations. {\bf First}, compared with GMN \cite{li2019graph}, SimGNN \cite{bai2019simgnn} and GraphSim \cite{bai2020learning}, our method obtains relative gains around 5\%. It indicates that our method makes better use of node-wise interactions. {\bf Second}, compared with PSimGNN \cite{xu2021graph} and H2MN \cite{zhang2021h2mn}, our method obtains relative gains of around 2\% . It implies that the improvement of graph representation ability can benefit experimental results. A more detailed analysis can be found in the ablation study. {\bf Notably}, we only exploit node-wise interactions at the last layer, while the previous method exploits the interaction information at each layer. It shows that our computational paradigm is simpler and has less complexity. Moreover, the construction process of the classification label is inconsistent with the internal logic of our method. It demonstrates that our method is robust even if the labels are MCS-independent.

{\bf Graph-Graph Regression Task }For the graph-graph regression task, we also conduct the experiments
five times and report their mean performance. The detailed performances on real and synthetic datasets are demonstrated in Table \ref{t_reg_mcs_real} and \ref{t_reg_ba}. The results for the GED metric can be found in the appendix. Our model achieves state-of-the-art performance on both MCS and GED metrics. In general, we can obtain similar conclusions as to the classification task. As for synthetic datasets, we observe that our method still outperforms other methods. It shows that our method can be adapted to bigger graphs. Notably, the results of our method on the MCS metric are about eight times better than those on the GED metric. We infer that this situation results from the model's internal logic consistent with the MCS label. The MSE of the model prediction results is close to zero, which means we can infer a more accurate MCS based on the average size of the input graph pairs. It increases the interpretability.

\begin{table}[]
\caption{Graph-Graph regression results about mse($\times 10^{-2}$), $\rho$ and p@10 on the MCS metric.}
\label{t_reg_mcs_real}
\centering
\begin{tabular}{c|ccc|ccc|ccc}
\hline
Datasets & \multicolumn{3}{c|}{AIDS(2-15)}                     & \multicolumn{3}{c|}{LINUX(2-15)}                                  & \multicolumn{3}{c}{PTC\_MM(all)}                                  \\ \hline
Metrics  & mse$\downarrow$ & $\rho$$\uparrow$  & p@10$\uparrow$  & mse$\downarrow$ & $\rho$$\uparrow$ & p@10$\uparrow$  & mse$\downarrow$ & $\rho$$\uparrow$ & p@10$\uparrow$  \\ \hline
\multicolumn{1}{l|}{EMBAVG}                        & 33.20           & 0.0045          & 0.0540          & 0.83            & 0.5922                        & 0.1340          & 35.03           & 0.0497                        & 0.3471          \\
\multicolumn{1}{l|}{GMN}                           & 32.20           & 0.0039          & 0.0578          & 3.99            & 0.0561                        & 0.1340          & 35.03           & 0.0370                        & 0.3500          \\
\multicolumn{1}{l|}{GraphSim}                      & 2.73            & 0.1688          & 0.0578          & 0.81            & 0.2260                        & 0.1340          & 3.21            & 0.5001                        & 0.3500          \\
\multicolumn{1}{l|}{SimGNN}                        & 2.65            & 0.1784          & 0.0596          & 0.83            & 0.4281                        & 0.2370           & 3.27            & 0.5280                        & 0.3500          \\
\multicolumn{1}{l|}{SMPNN}                         & 2.89            & 0.2046          & 0.1056          & 12.59           & 0.5502                        & 0.4280          & 4.67            & 0.4558                        & 0.4353          \\
\multicolumn{1}{l|}{MGMN}                          & 1.69            & 0.5300          & 0.1683          & 0.87            & 0.5351                        & 0.3664          & 1.43            & 0.7329                        & 0.5200          \\
\multicolumn{1}{l|}{PSimGNN}                       & 2.54            & 0.1031          & 0.0452          & 1.83            & 0.4311                        & 0.2668          & 3.43            & 0.4359                        & 0.4280          \\
\multicolumn{1}{l|}{GOTSim}                       & 1.77            & 0.5550          & 0.1763          & 0.61            & 0.3752                        & 0.2569          & 2.75            & 0.3495                        & 0.3431          \\
\multicolumn{1}{l|}{H2MN}                          & 1.29           & 0.6745          & 0.2097          & 0.44            & 0.6364                        & 0.4795          & 1.07            & 0.8823                        & 0.7182          \\ \hline
\multicolumn{1}{l|}{INFMCS}                       & \textbf{0.30}   & \textbf{0.9352} & \textbf{0.7976} & \textbf{0.02}   & \textbf{0.9814}               & \textbf{0.8870} & \textbf{0.71}   & \textbf{0.9205}               & \textbf{0.7794} \\ \hline
\end{tabular}
\end{table}

\begin{table}[]
\caption{Graph-Graph regression results about mse($\times 10^{-2}$) on the synthetic datasets.}
\label{t_reg_ba}
\centering
\begin{tabular}{c|cc|cc|cc}
\hline
Datasets & \multicolumn{2}{c|}{BA100}        & \multicolumn{2}{c|}{BA200}        & \multicolumn{2}{c}{BA300}      \\ \hline
Metric   & mse(MCS)         & mse(GED)       & mse(MCS)         & mse(GED)       & mse(MCS)      & mse(GED)       \\ \hline
\multicolumn{1}{l|}{EMBAVG}    & 16.21            & 10.581         & 20.24            & 9.171          & 21.79         & 12.732         \\
\multicolumn{1}{l|}{GMN}       & 16.21            & 8.831          & 20.24            & 9.002          & 20.14         & 8.756          \\
\multicolumn{1}{l|}{
 GraphSim} & 0.20             & 0.065          & 0.44             & 0.140          & 0.57          & 0.062          \\
\multicolumn{1}{l|}{SimGNN}    & 0.20             & \textbf{0.060}          & 0.05             & 0.180          & 0.02          & 0.110          \\
\multicolumn{1}{l|}{SMPNN}     & 1.10             & 22.530         & 0.32             & 23.920         & 0.24          & 24.290         \\
\multicolumn{1}{l|}{MGMN}      & 0.35             & 1.033            & 0.27             & 0.901            &    0.44         & 0.071            \\
\multicolumn{1}{l|}{PSimGNN}   & 0.48             & 1.932            & 0.51             & 1.366            &    0.67         & 0.103            \\
\multicolumn{1}{l|}{H2MN}      & 0.02             & 0.187            & 0.01             & 0.532            & 0.02             & 0.034            \\ \hline
\multicolumn{1}{l|}{INFMCS}  & \textbf{5.49e-7} & \textbf{0.061} & \textbf{1.80e-5} & \textbf{0.011} & \textbf{0.003} & \textbf{0.005} \\ \hline
\end{tabular}
\end{table}

\begin{figure}
	\centering
	\begin{minipage}[t]{0.4\textwidth}
	\centering
    	\subfigure{
    		\begin{minipage}[b]{1\textwidth}
    		\centering
  		 	\centerline{\includegraphics[width=1\textwidth]{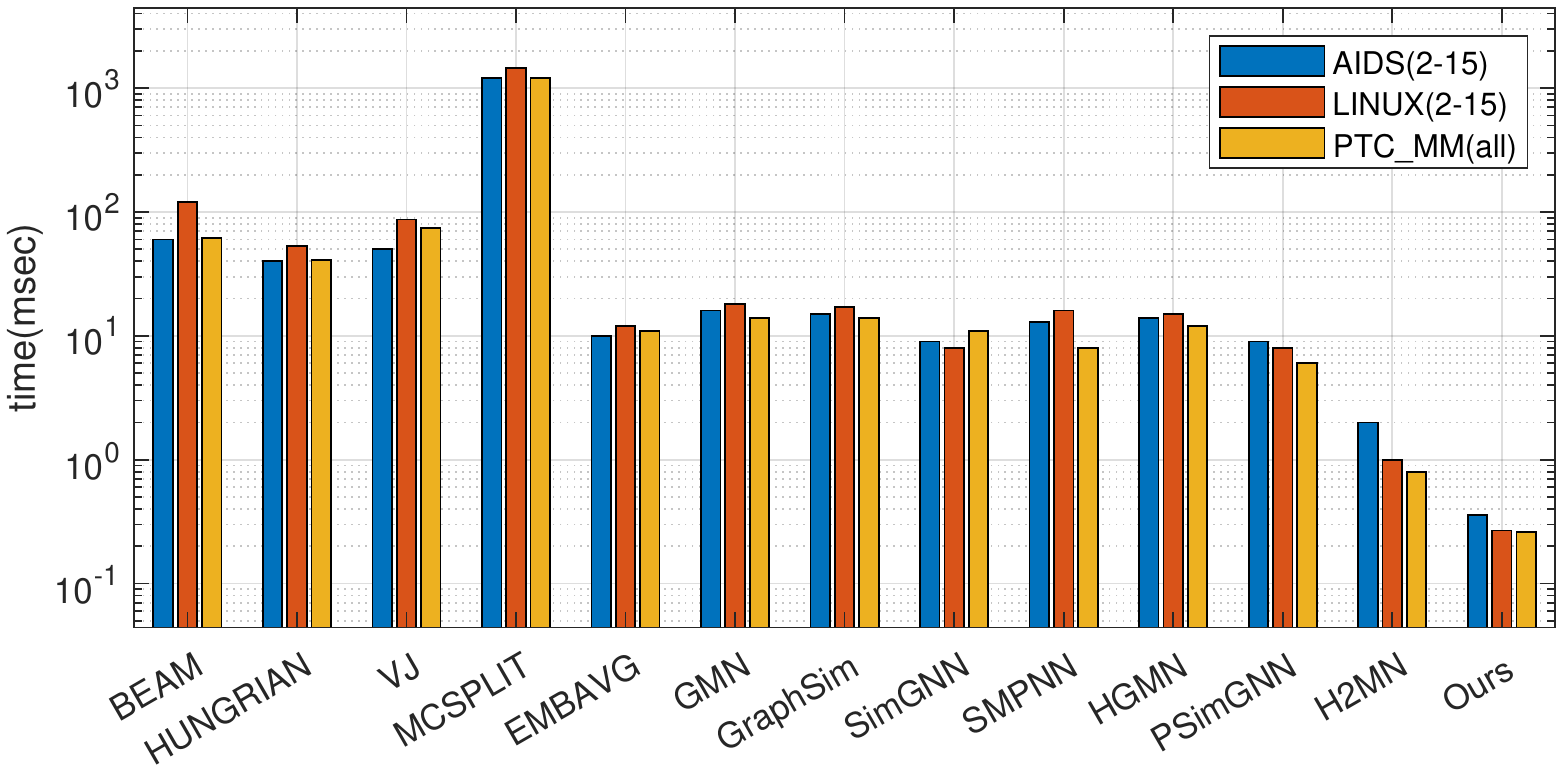}}
  		 	\caption{Running time comparisons.}
  		 	\label{f_efficiency}
  		 	
  		 	\centering
		 	\centerline{\includegraphics[width=1\textwidth]{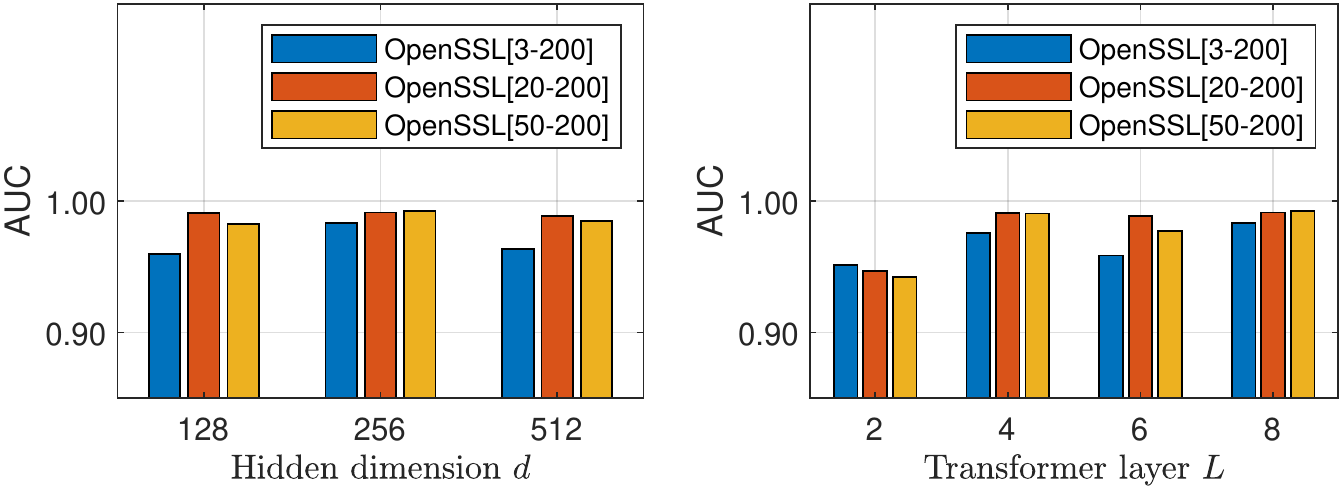}}
		 	\caption{Sensitive analysis of test set.}
		 	\label{f_sensitivity}
    		\end{minipage}
    	}
	\end{minipage} \,\,\,\,\,\,\,\,\,\,\,\,\,\,\,\,\
	\begin{minipage}[t]{0.5\textwidth}
	\centering
	\subfigure{
		\begin{minipage}[b]{0.95\textwidth}
		\centering
		
			\centerline{\includegraphics[width=1\textwidth]{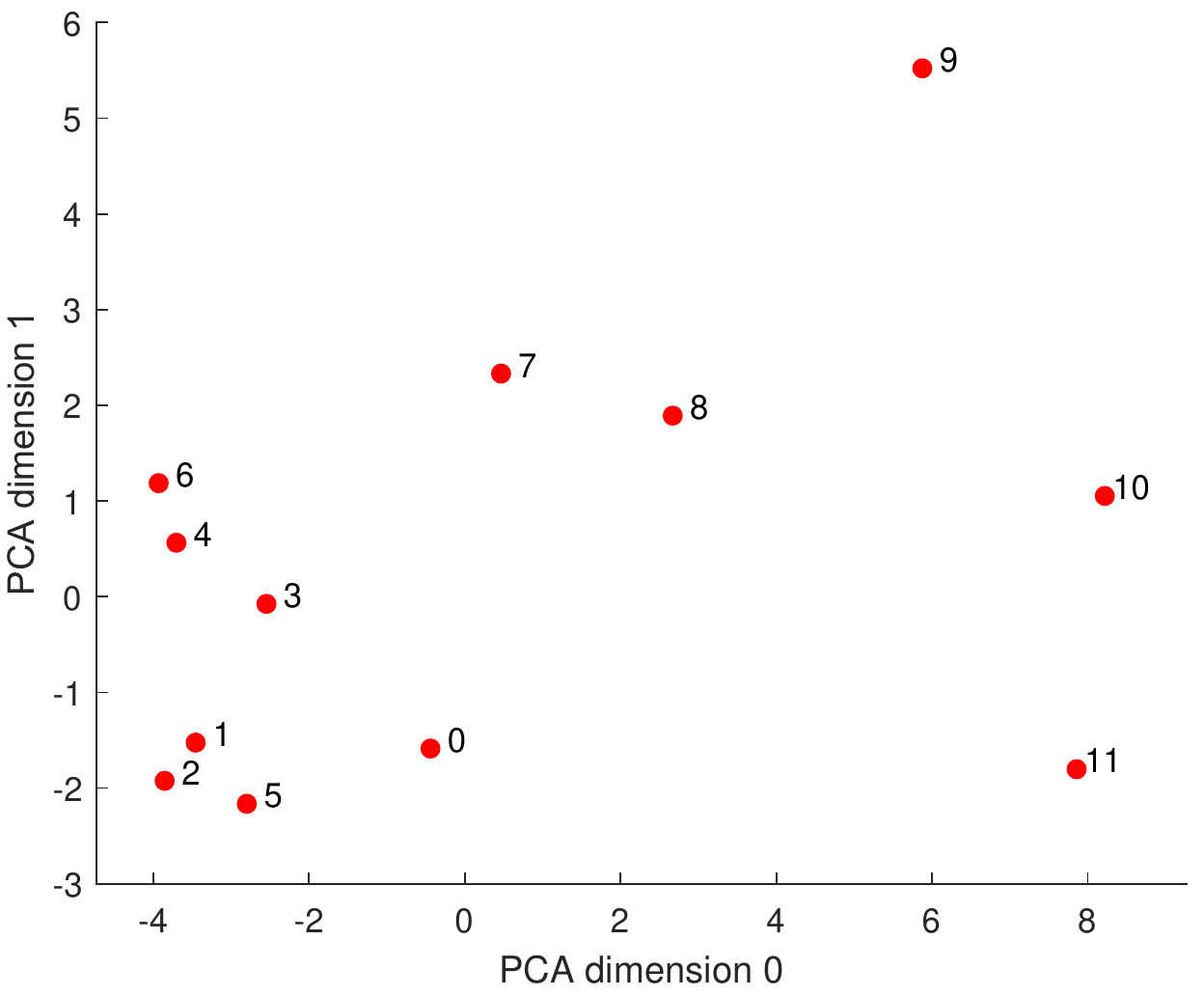}} 
			\caption{Positional Encoding analysis.}
			\label{f_PE}
		\end{minipage}
	}
	\end{minipage}
\end{figure}

{\bf Efficiency }We show the running time between different methods in Figure \ref{f_efficiency} in order to evaluate the efficiency of INFMCS. As we can see, the learning-based approaches are consistently faster than the traditional methods in all datasets, especially the exact algorithm MCSPLIT. We also observe that INFMCS is faster than other learning-based approaches. We attribute the efficiency gains to two points. First, unlike previous methods that exploit the interaction information of each layer, our method only needs to compute the interaction information of the last layer. Second, our method does not require building hypergraphs or graph partitions.

{\bf Hyperparameter sensitivity analysis }We fixed the number of Transformer layers $L$ to 8 and the dimension of hidden layers $d$ to 256 to explore the impact of several vital hyper-parameters on OpenSSL subsets (Figure \ref{f_sensitivity}). The performance under different hyperparameters is consistent, revealing the robustness of our method. We observe that the performance of INFMCS improves as the number of Transformer layers increases. We hypothesise that more Transformer layers allow node-level embeddings to contain more global information, thus improving experimental results. A comparison of the number of parameters of our method with other baselines can be found in appendix.

\subsection{Ablation Study}
{\bf BASE}\footnote{MCS-RSC is MCS-Related Similarity Computation. BASE = GCN + MCS-RSC; BASE+T = GCwT-w/o-PE + MCS-RSC; BASE+T+PE = GCwT-w/-PE + MCS-RSC; BASE+H = HyperGCN + MCS-RSC; H2MN-H = GCN + H2MN-RSC.} denotes the proposed similarity computation paradigm whose graph representation model (GRM) is GCN. {\bf BASE+T}'s GRM is GCwT without the permutation-invariant Positional Encoding. {\bf BASE+T+PE}'s GRM is GCwT with the permutation-invariant Positional Encoding. {\bf BASE+}$\mathbf{H}_{rw}$'s GRM is hypergraph convolution used in \cite{zhang2021h2mn}. We use the part of their code\footnote{https://github.com/cszhangzhen/H2MN} to obtain node embeddings and then pass these to our paradigm end-to-end\footnote{We use the random walk to construct the hypergraph and set hyperparameter k to 5}. {\bf H2MN-H} denotes that the H2MN model's hyperparameter k is set to 1, indicating that hypergraph convolution degenerates to GCN. The results of the ablation study is illustrated in Table \ref{t_ab_auc} and \ref{t_ab_mcs}, and on which we have the four observations: {\bf 1)} {\bf BASE} outperforms previous methods except for H2MN, thus we compare {\bf BASE} with {\bf H2MN-H} in order to demonstrate the effectiveness of the computation paradigm. We find {\bf BASE} outperforms {\bf H2MN-H}, which means more interpretable forward computation can improve performance. {\bf 2)} {\bf BASE+T} has lower performance than {\bf BASE}. It indicates that the performance of GCwT without the permutation-invariant Positional Encoding degrades. We attribute the reason to the loss of graph structural information since stacking Transformer layers only is equivalent to treating the graph as a fully connected graph. {\bf 3)} {\bf BASE+T+PE} outperforms {\bf BASE}, which confirms that GCwT with the permutation-invariant Positional Encoding not only increases the receptive field of the node but also learns the structural information through PE. More on Positional Encoding analysis can be found in the next subsection. {\bf
4)} {\bf BASE+T+PE} outperforms {\bf BASE+}$\mathbf{H}_{rw}$, implying that GCwT can be better adapted to the proposed computation paradigm and have decent global representation.

\begin{minipage}{\textwidth}
    \begin{minipage}[t]{0.5\textwidth}
    \captionof{table}{Ablation study on the FFmpeg.}
    \label{t_ab_auc}
    \centering
    \begin{tabular}{l|ccc}
    \hline
    \multicolumn{1}{c|}{(AUC score)} & 3-200 & 20-200 & 50-200 \\ \hline
    H2MN-H                        & 97.50       & 98.12        & 98.05         \\
    BASE                          & 98.16       & 98.83        & 98.87         \\
    BASE+T                        & 97.13       & 98.20        & 98.48         \\
    BASE+H                        & 98.01            &98.42   & 98.56             \\
    BASE+T+PE                     & 98.49       & 99.36        & 99.49         \\ \hline
    \end{tabular}
    \end{minipage}
    \begin{minipage}[t]{0.5\textwidth}
    \captionof{table}{Ablation study on the MCS metric.}
    \label{t_ab_mcs}
    \centering
    \begin{tabular}{l|ccc}
    \hline
    \multicolumn{1}{c|}{(mse$\times 10^{-2}$)} & AIDS & LINUX & PTC\_MM \\ \hline
    H2MN-H                        & 1.63       & 0.56        & 1.18         \\
    BASE                          & 1.41       & 0.36        & 0.98         \\
    BASE+T                        & 3.21       & 0.93        & 1.24         \\
    BASE+H                        & 1.70           & 0.21        & 1.02             \\
    BASE+T+PE                     & 0.30       & 0.02        & 0.71         \\ \hline
    \end{tabular}
    \end{minipage}
\end{minipage}

\subsection{Inteprtability Analysis, Case Study and Discussion}

{\bf Positional Encoding }We reduce the Positional Encoding trained on AIDS(2-15) to two dimensions via PCA and present them in Fig. \ref{f_PE}, where each red point denotes a Positional Encoding. Since the size of most samples is no more than 12, embeddings after position 12 are insufficiently trained. We present the top 12 Positional Encoding embeddings and obtain an exciting observation. The Euclidean distance between Positional Encoding 0 and the following Positional Encoding gradually increases. The Euclidean distance here corresponds to each node's centrality in the graph, which means the Positional Encoding embeddings preserve the graph's structural information well.


\begin{figure}[tp]
\centering
\begin{minipage}[t]{0.49\linewidth}
\centering
    \includegraphics[width=0.9\linewidth]{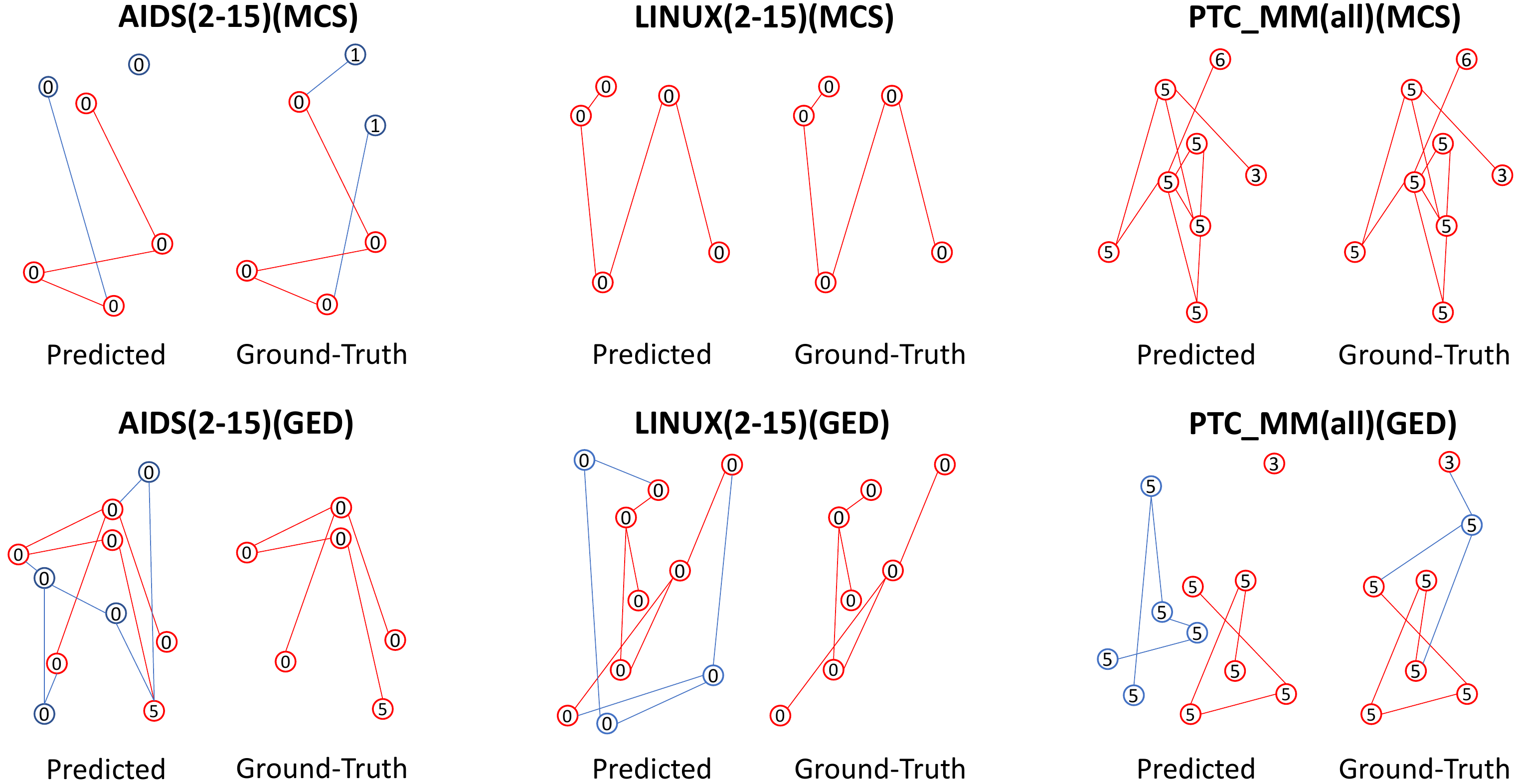}
    \captionof{figure}{Visualizations of inferring MCS.}
    \label{f_vis}
\end{minipage}%
   \,\,\,\ \hfill%
\begin{minipage}[t]{0.49\linewidth}
\centering
    \includegraphics[width=0.9\linewidth]{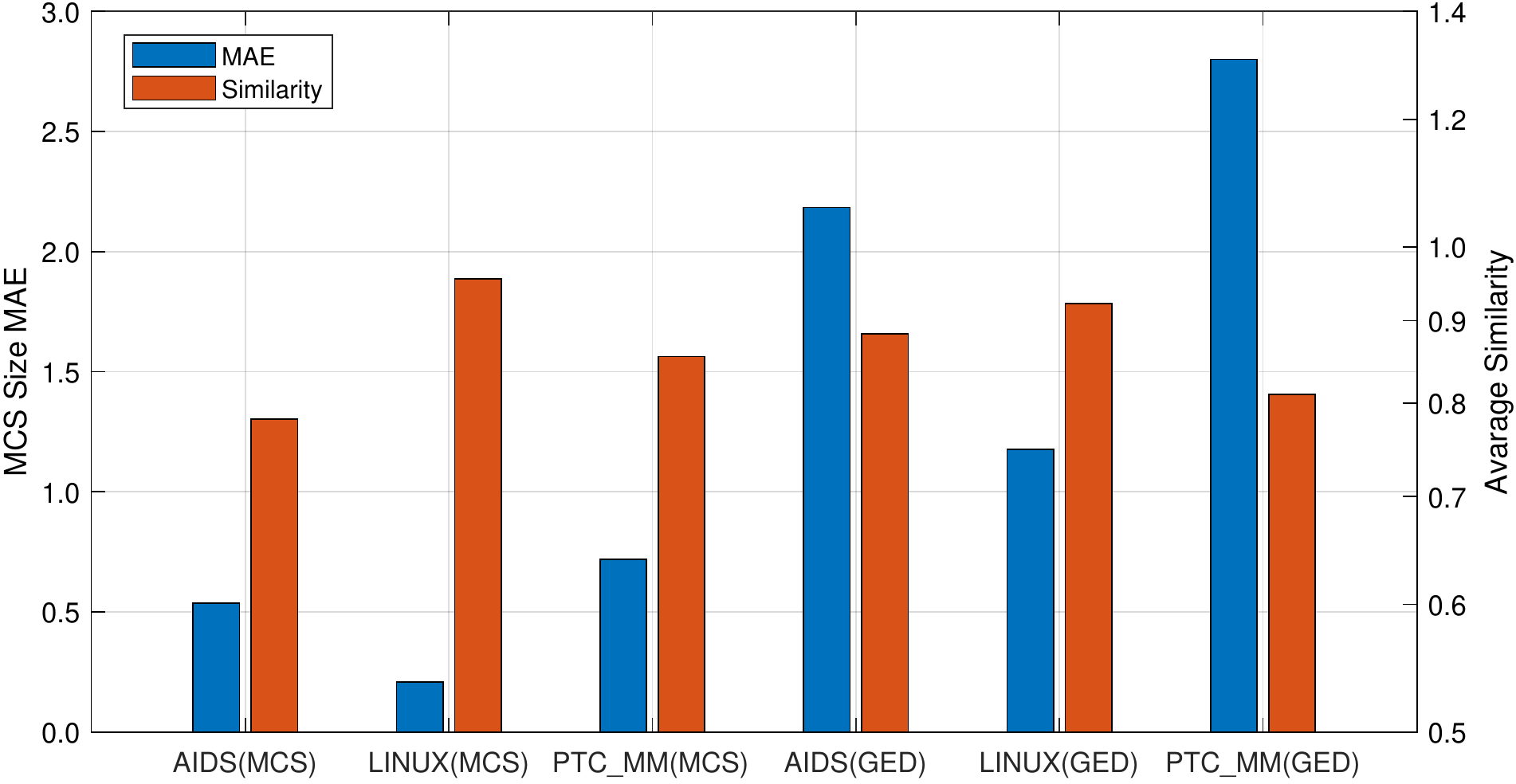}
    \caption{Interpretability analysis.}
    \label{f_inter}
\end{minipage} 
\end{figure}

\begin{figure}
	\centering
	\includegraphics[width=0.94\linewidth]{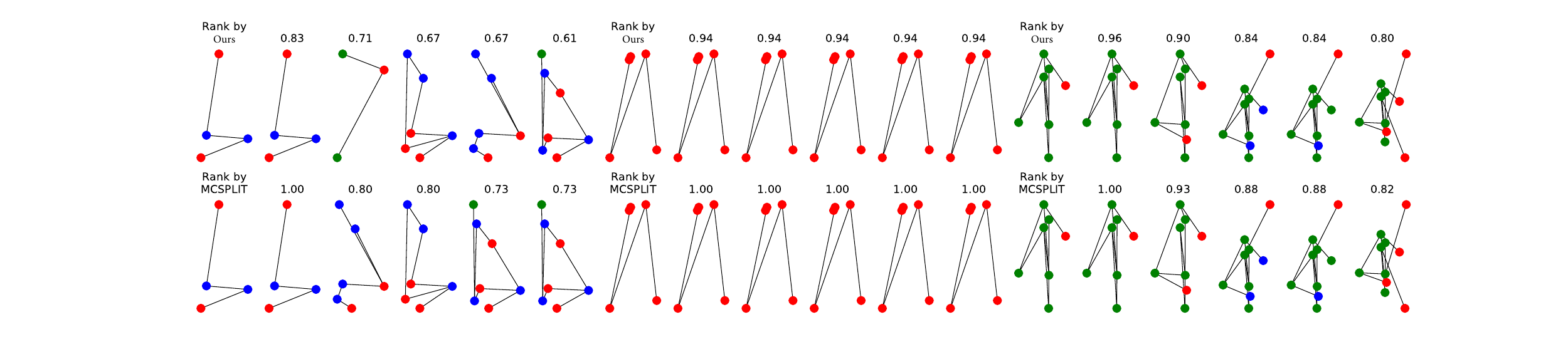}
	\caption{Visualization of ranking results. From left to right: AIDS, LINUX, PTC\_MM.}
	\label{f_ranks}
\end{figure}

{\bf Infer MCS }During inference, we infer the MCS size $m$ via multiplying the average size of the graph pair by the similarity score. Next, we extract a subgraph from $\mathcal{G}_{1}$ that consists of nodes corresponding to the top $m$ matching scores $s$ in Eq. \ref{e5}. Also, we use MCSPLIT to extract the true MCS. To evaluate the quality of our predicted MCS, we compute the similarity between the predicted and the actual MCS. The similarity results and some visualizations are presented in Fig. \ref{f_vis} and \ref{f_inter}, where red subgraphs are MCS between the predicted and the actual MCS. The mean absolute error between the predicted and actual sizes is larger on the GED metric. We attribute the reason to the inconsistency in the model logic and the label construction process. However, we note that the similarity between the predicted MCS and the ground-truth MCS is higher than 0.8 on both metrics except for the AIDS (MCS) results, revealing the interpretability of our approach. We can predict the similarity score and infer MCS to understand why this similarity score is predicted.

{\bf Case Study }We demonstrate three example queries, one from each dataset in Fig. \ref{f_ranks}. The first row shows the graphs returned by our model in each demo, with the predicted similarity for each graph shown at the top. The bottom row describes the graphs returned by MCSPLIT. Notably, the top 5 results are precisely the isomorphic graphs to the query in the case of LINUX and PTM\_MM.

{\bf Why is INFMCS effective? }For the computation paradigm, our method ultimately only needs to consider $n$ pairs of interaction information, which means it captures more critical information that affects the similarity compared to previous methods. Since Positional Encoding well preserves the positional information of nodes, the interaction within the model compares the similarity of local structures between nodes and considers the relative position between nodes in two graphs. Intuitively, the more similar the two graphs are, the more similar their corresponding local regions should be.

\section{Conclusion}
This paper proposes a more interpretable end-to-end paradigm for graph similarity learning, whose interpretable computation process improves the performance of graph similarity learning. The model can implicitly infer Maximum Common Subgraph during inference. We stack some vanilla transformer encoder layers with graph convolution layers and propose a novel permutation-invariant node Positional Encoding to capture more global information. Comprehensive experiments and ablation studies demonstrate that INFMCS outperforms previous methods and is more interpretable.

\appendix

\section{Graph Convolution Layer}
We describe the computation process of one layer in $\mathrm{GCN}(\cdot)$:
\begin{equation}
\label{gcn}
u_{i}^{n+1}=\operatorname{ReLU}\left(\sum_{j \in \mathcal{N}(i)} \frac{1}{\sqrt{d_{i} d_{j}}} u_{j}^{n} \boldsymbol{W}^{(n)}+b^{(n)}\right).
\end{equation}

Here, $u_{i}^{(n)} \in \mathbb{R}^{D^{(n)}}$ and $u_{i}^{(n+1)} \in \mathbb{R}^{D^ {(n+1)}}$ are representations of node $i$ at $n$-th and $n+1$-th layer. $\mathcal{N}(i)$ is the set of the first-order neighbors of node $i$ plus $i$ itself. $d_{i}$ is the degree of node $i$ plus 1. $\boldsymbol{W}^{(n)} \in \mathbb{R}^{D^{(n)} \times D^{(n+1)}}$ is the weight matrix of the $n$-th GCN layer. $\boldsymbol{b}^{(n)} \in \mathbb{R}^{D^{(n+1)}}$ is the bias, $D^{(n)}$ denotes the dimension of embedding vector at layer $n$. $\operatorname{ReLU}(x)=\max (0, x)$ is the activation function.

\section{Transformer Encoder Layer}
Each Transformer Encoder Layer has two parts: a self-attention module and a position-wise feed-forward network ($\operatorname{FFN}$):
\begin{equation}
Q^{h}=H^{(l)} W_{Q}^{h}, \quad K^{h}=H^{(l)} W_{K}^{h}, \quad V^{h}=H^{(l)} W_{V}^{h},
\end{equation}
\begin{equation}
A^{h}=\frac{Q^{h} {K^{h}}^{\top}}{\sqrt{d_{K}}}, H^{h}=\operatorname{softmax}(A^{h}) V^{h},
\end{equation}
\begin{equation}
H^{\prime} = W \cdot \left(\|_{h} H^{h}\right) + H^{(l)}
\end{equation}
\begin{equation}
H^{(l+1)}=\operatorname{FFN}\left(\mathrm{LN}\left(H^{\prime}\right)\right)+H^{\prime}.
\end{equation}

Here, $H^{(l)}=\left[h_{1}^{(l) \top}, \cdots, h_{n}^{(l) \top}\right]^{\top} \in \mathbb{R}^{n \times d}$ denote the input of self-attention module where $d$ is the hidden dimension and $h_{i}^{(l)} \in \mathbb{R}^{1 \times d}$ is the hidden representation at position $i$. The input $H$ is projected by three matrices $W_{Q}^{h} \in \mathbb{R}^{d \times d_{K}}$, $W_{K}^{h} \in \mathbb{R}^{d \times d_{K}}$ and $W_{V}^{h} \in \mathbb{R}^{d \times d_{V}}$ to the corresponding representations $W_{Q}^{h}$, $W_{K}^{h}$, $W_{V}^{h}$ for each head $h$. $A$ is a matrix capturing the similarity between queries and keys for each head $h$ and $\|$ is concatenation operation. $\mathrm{LN}$ is layer normalization and $\operatorname{FFN}$ is the feed-forward network. For simplicity, we omit bias terms and assume $d_{K}=d_{V}=d$.

\section{Datasets and Preprocessing}
\subsection{Real Datasets}
For {\bf AIDS(2-15)}, {\bf LINUX(2-15)}, we randomly selected 1500 graphs whose size range from 2 to 15 in the {\bf AIDS} \cite{riesen2008iam} and {\bf LINUX} \cite{wang2012efficient}. For {\bf PTC\_MM(all)} , we use the all graphs in the {\bf PTC\_MM} \cite{helma2001predictive}. Each dataset is randomly split 80\%, 10\%, and 10\% of all the graphs as original training set, original validation set, and original testing set, respectively. Then, each graph in the original set is paired with another graph as a sample. The datasets statistics for the regression task is shown in Table. \ref{st_real}. 

\makeatletter\def\@captype{table}\makeatother
\begin{center}
\caption{Statistics of AIDS, LINUX and PTC\_MM. B (Billion), M (Million).}
\label{st_real}
\centering
\begin{tabular}{l|cccc}
\hline
\multicolumn{1}{c|}{} & $\# |G|$ & $Avg. (|V|, |E|)$ & \multicolumn{1}{l}{\#Labels} & \# Graph Pairs \\ \hline
AIDS(2-15)            & 1500     & (10.22, 20.39)    & 38                           & 225M        \\
LINUX(2-15)           & 1500     & (9.40, 15.58)     & 1                            & 225M       \\
PTC\_MM(all)          & 336      & (13.97, 28.64)    & 20                           & \textasciitilde 11.2M         \\ \hline
\end{tabular}
\end{center}

For MCS metric, we use MCSPLIT \cite{mccreesh2017partitioning} to caculate the MCS for a pair of graphs and then normalize it to similarity score: $\mathrm{nMCS}\left(\mathcal{G}_{1}, \mathcal{G}_{2}\right)= \frac{\left|\mathrm{MCS}\left(\mathcal{G}_{1},\mathcal{G}_{2}\right)\right|}{\left(\left|\mathcal{G}_{1}\right|+\left|\mathcal{G}_{2}\right|\right) / 2}$. For GED metric, we use A* \cite{riesen2013novel} to compute GED for graphs. For graphs that can not be computed by A* in 1000 seconds, we take the minimum distance computed by BEAM \cite{neuhaus2006fast}, HUNGARIAN \cite{riesen2009approximate}, VJ \cite{fankhauser2011speeding} and HED \cite{fischer2015approximation}, since their returned GEDs are guaranteed to be upper bounds of the true GEDs. Figure. \ref{pie} presents the proportion of exact and approximate GED-labels on the three datasets. Next, we normalize it to similarity score: $\mathrm{nGED}\left(\mathcal{G}_{1}, \mathcal{G}_{2}\right)=\exp(\frac{-\mathrm{GED}\left(\mathcal{G}_{1}, \mathcal{G}_{2}\right)}{\left(\left|\mathcal{G}_{1}\right|+\left|\mathcal{G}_{2}\right|\right) / 2})$. Figure. \ref{distribution} presents the distribution of the similarity score on two metric.

\makeatletter\def\@captype{figure}\makeatother
\begin{center}
	\centering
	\includegraphics[width=1\linewidth]{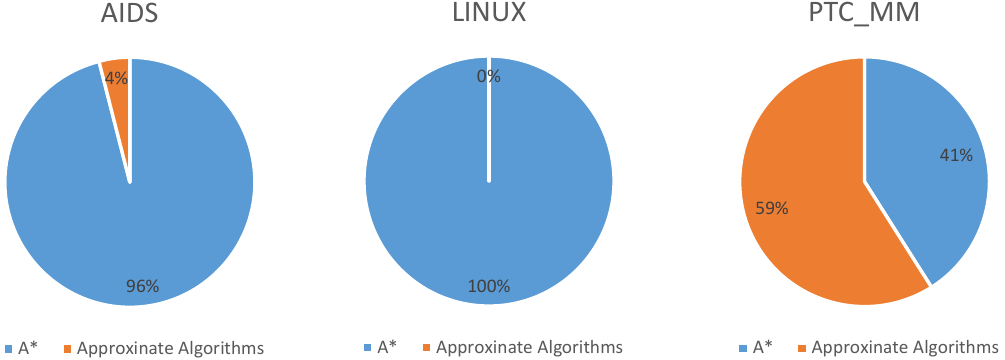}
	\caption{The  proportion of exact and approximate GED-labels.}
	\label{pie}
\end{center}

\makeatletter\def\@captype{figure}\makeatother
\begin{center}
	\centering
	\includegraphics[width=1\linewidth]{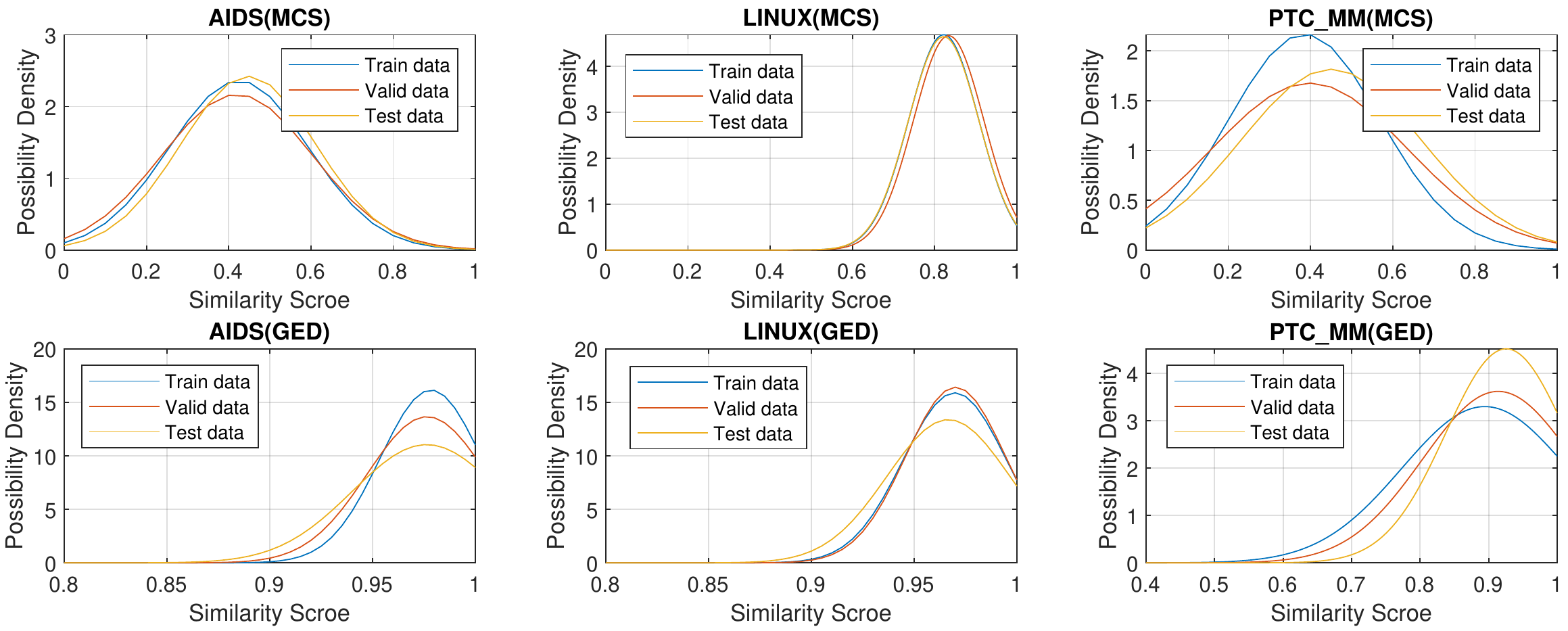}
	\caption{The distribution of the real datasets.}
	\label{distribution}
\end{center}

For the classification task, we split \textbf{FFmepeg} and \textbf{OpenSSL} into 3 sub-datasets (i.e., [3, 200], [20,200], and [50,200]) according to the size ranges of pairs of input graphs. Table. \ref{st_cls} shows the statistics of these datasets. 

\makeatletter\def\@captype{table}\makeatother
\begin{center}
\caption{Statistic of FFmpeg and OpenSSL. B (Billion), M (Million).}
\label{st_cls}
\centering
\begin{tabular}{l|ccccc}
\hline
\multicolumn{1}{c|}{}    & Subsets       & $\# |G|$ & $Avg. |V|$ & $Avg. |E|$ & \# Graph Pairs \\ \hline
\multirow{3}{*}{FFmpeg}  & {[}3, 200{]}  & 83008    & 18.83      & 27.02      & 6.89B          \\
                         & {[}20, 200{]} & 31696    & 51.02      & 75.88      & 1B             \\
                         & {[}50, 200{]} & 10824    & 90.93      & 136.83     & 117M           \\ \hline
\multirow{3}{*}{OpenSSL} & {[}3, 200{]}  & 73953    & 15.73      & 21.97      & 5.46B          \\
                         & {[}20, 200{]} & 15800    & 44.89      & 67.15      & 249M           \\
                         & {[}50, 200{]} & 4308     & 83.68      & 127.75     & 18.5M          \\ \hline
\end{tabular}
\end{center}

\subsection{Synthetic Datasets}
To verify the performance on large graphs, we also use the Barabási–Albert model \cite{jeong2003measuring} to generate synthetic graphs. We generated three sythetic datasets with graph sizes around 100, 200, and 300 respectively, called {\bf BA100}, {\bf BA200} and {\bf BA300}. 

{\bf MCS metric} We randomly generate a core graph by BA-model for each example and take this core as the primary graph. Next, we add edges and nodes to its surroundings randomly, thereby obtaining ${g}_{1}$ and ${g}_{2}$. We regard the core as the MCS and then obtain the similarity score. For instance, the size of the core graph is 60, and the number of added nodes for ${g}_{1}$ is 35, for ${g}_{2}$ is 45, thus the final label is 0.6. Although the size of the real MCS may be more than 60, it is the lower bound for this sample. The range of core graph size and added nodes are shown in Table \ref{sts_ba_mcs}. The pseudocode is shown in the Alg. \ref{mcs_algorithm}. 

\makeatletter\def\@captype{table}\makeatother
\begin{center}
\centering
\caption{Statistics of BA-graph for MCS metric.}
\label{sts_ba_mcs}
\begin{tabular}{c|ccccc}
\hline
       & \# Size of Core     & \# Added Nodes     & \# Train Pairs & \# Valid Pairs & \# Test Pairs \\ \hline
BA100 & (50, 70)   & (30, 50)   & 32000          & 4000           & 4000         \\
BA200 & (100, 120) & (80, 100)  & 32000          & 4000           & 4000         \\
BA300 & (150, 170) & (130, 150) & 32000          & 4000           & 4000         \\ \hline
\end{tabular}
\end{center}

\makeatletter\def\@captype{algorithm}\makeatother
\begin{algorithm}
\caption{Generate BA graphs for MCS metric}\label{mcs_algorithm}
\KwIn{$(c_1, c_2)$ for the range of the number of core graph nodes; $(a_1, a_2)$ for the range of the number of added graph nodes; $n$ for the number of samples}
\KwOut{BA datasets in MCS metric}
\BlankLine
$data\leftarrow \{ \}$\;
\For{$i\leftarrow 0$ \KwTo $n$}{
    $c\leftarrow random(c_1, c_2)$\;
    $\mathrm{a}_{1}\leftarrow random(a_1, a_2)$\;
    $\mathrm{a}_{2}\leftarrow random(a_1, a_2)$\;
    \tcp{use BA-model to generate the core graph with $c$ nodes}
    $core\leftarrow barabasi\_albert\_graph(c)$\;
    \tcp{use BA-model to generate two graphs based on the core graph, and they have $c+a$ nodes}
    $g_1^i\leftarrow barabasi\_albert\_graph(c+\mathrm{a}_{1}, core)$\;
    $g_2^i\leftarrow barabasi\_albert\_graph(c+\mathrm{a}_{2}, core)$\;
    $sim\leftarrow c / (c + 0.5 \times (\mathrm{a}_{1} + \mathrm{a}_{2}))$\;
    $\{g_1^i, g_2^i, sim\} || data$\;
}   
\end{algorithm}

{\bf GED metric} Firstly, we generated two basic graphs by BA-model. To each of them, it is edited by \textbf{1.} Deleting a leaf node and its edge \textbf{2.} Adding a leaf node \textbf{3.} Adding an edge among existing node. The trimming step $n$ from the basic graph $g$ to the generated one $g_{i}$ ranges from 1 to 10, which means the GED score between $g$ and $g_{i}$ ranges from 1 to 10. In order to get more data, we combine two generated graphs and derive the GED by adding their edit distances coming from the basic graph ($n_{1} + n_{2}$), If the GED computed by BEAM \cite{neuhaus2006fast}, HUNGARIAN \cite{riesen2009approximate}, VJ \cite{fankhauser2011speeding} and HED \cite{fischer2015approximation} is less than $n_{1} + n_{2}$, we use the GED computed by the approximate algorithm.The datasets statistics is shown in Table. \ref{sts_ba_ged}. The pseudocode is shown in the Alg. \ref{ged_algorithm}.

\makeatletter\def\@captype{table}\makeatother
\begin{center}
\centering
\caption{Statistics of BA-graph for GED metric.}
\label{sts_ba_ged}
\begin{tabular}{c|ccccc}
\hline
      & \# Basic Graph & \# Nodes & \# Train Pairs & \# Valid Pairs & \# Test Pairs \\ \hline
BA100 & 2            & 100                   & 32000          & 4000           & 4000         \\
BA200 & 2            & 200                   & 32000          & 4000           & 4000         \\
BA300 & 2            & 300                   & 32000          & 4000           & 4000         \\ \hline
\end{tabular}
\end{center}

\makeatletter\def\@captype{algorithm}\makeatother
\begin{algorithm}
\caption{Generate BA graphs for GED metric}\label{ged_algorithm}
\KwIn{$b$ for the number of basic graph nodes; $n$ for the number of generated samples to store in one collection}
\KwOut{BA datasets in GED metric}
\BlankLine
$data\leftarrow \{ \}$\;
$ged\leftarrow 0$\;
$A\leftarrow \{ \}$\;
$B\leftarrow \{ \}$\;
\BlankLine
Generate two collections of graphs\;
\For{$i\leftarrow 0$ \KwTo $n$}{
    \tcp{use BA-model to generate two basic graphs with $b$ nodes}
    $b_1\leftarrow barabasi\_albert\_graph(b)$\;
    $b_2\leftarrow barabasi\_albert\_graph(b)$\;
    \tcp{every 10 iterations, increase ged}
    \If{$i\ \%\ 10 = 0$}{
        $ged\leftarrow ged + 1$\;
    }
    \tcp{the trimming methods are 1.deleting a leaf node and its edge 2.adding a leaf node and 3.adding an edge, which have the same possibility to operate for each time}
    $g_1\leftarrow trim\_ged\_times(b_1)$\;
    $g_2\leftarrow trim\_ged\_times(b_2)$\;
    $\{g_1, ged\} || A$\;
    $\{g_2, ged\} || B$\;
}   
\BlankLine
Pack and label graphs\;
$C\leftarrow A||B$\;
\For{$i\leftarrow 0$ \KwTo $2n$}{
    \For{$j\leftarrow 0$ \KwTo $2n$}{
        $(g_1, ged_1)\leftarrow C[i]$\;
        $(g_2, ged_2)\leftarrow C[j]$\;
        $ged\leftarrow min(ged_1 + ged_2, min($BEAM$, $HUNGRIAN$, $VJ$, $HED$))$\;
        $sim\leftarrow \exp(\frac{-ged}{|g_1| + |g_2|) / 2})$\;
        $\{g_1, g_2, sim\} || data$\;
    }
}

\end{algorithm}

\section{The Results for GED Metric}
The regression results on the GED metric can be found in Table. \ref{ged_rst}.

\begin{table}[]
\centering
\caption{The regression results for the GED metric (mse $\times 10^{-2}$).}
\label{ged_rst}
\begin{tabular}{l|ccc|ccc|ccc}
\hline
\multicolumn{1}{c|}{}        & \multicolumn{3}{c|}{AIDS(2-15)}                                                                         & \multicolumn{3}{c|}{LINUX(2-15)}                                                                        & \multicolumn{3}{c}{PTC\_MM(all)}                                                                        \\ \hline
\multicolumn{1}{c|}{Metrics} & mse$\downarrow$ & $\rho$$\uparrow$  & p@10$\uparrow$  & mse$\downarrow$ & $\rho$$\uparrow$ & p@10$\uparrow$  & mse$\downarrow$ & $\rho$$\uparrow$ & p@10$\uparrow$  \\ \hline
GMN                          & 0.19                         & 0.0918                                     & 0.1857                      & 0.20                         & 0.0662                                     & 0.1400                      & 1.33                         & 0.0117                                     & 0.3324                      \\
GraphSim                     & 0.09                         & 0.3513                                     & 0.0641                      & 0.10                         & 0.1457                                     & 0.1400                      & 0.98                         & 0.1714                                     & 0.3382                      \\
SimGNN                       & 0.07                         & 0.4056                                     & 0.1342                      & 0.11                         & 0.4006                                     & 0.0810                      & 1.18                         & 0.1451                                     & 0.3382                      \\
SMPNN                        & 15.61                        & 0.1583                                     & 0.0634                      & 21.95                        & 0.0422                                     & 0.055                       & 21.27                        & 0.0806                                     & 0.3088                      \\
MGMN                         & 0.08                         & 0.2905                                     & 0.0437                      & 0.08                         & 0.2192                                     & 0.0106                      & 1.17                         & 0.1630                                     & 0.0891                      \\
PSimGNN                      & 0.09                         & 0.3361                                     & 0.0534                      & 0.11                         & 0.1803                                     & 0.0093                      & 1.21                         & 0.1520                                     & 0.0437                      \\
H2MN                         & 0.07                         & 0.3988                                     & 0.0540                      & 0.07                         & 0.3813                                     & 0.2763                      & 0.96                         & 0.1531                                     & 0.0990                      \\ \hline
INFMCS                       & \bf 0.04                         & \bf 0.4090                                     & \bf 0.1854                      & \bf 0.04                         & \bf 0.4054                                     & \bf 0.3445                      & \bf 0.68                         & \bf 0.1768                                     & \bf 0.3406   \\ \hline                   
\end{tabular}
\end{table}

\section{Hyperparameter Settings}
For all datasets, the GCN learning layer is fixed to 3, and the transformer encoder head number is set to 8; The scope of the hyperparameter search is shown in the Table. \ref{scope_p}. The results of the hyperparameter search are shown in Table. \ref{hyperparameter}.

\section{Complexity Analysis}
In this section, we analyze the time complexity of our proposed INFMCS. Here, $l_{1}$ denotes the number of GCN layers, $l_{2}$ denotes the number of Transformer Encoder layers, $d$ represents the hidden dimension and $h$ is nunber of heads in Transformer. First, the complexity of GCN is $\mathcal{O}(l_{1}|\mathcal{E}||\mathcal{V}|d^2)$. Second, closeness centrality requires the time of $\mathcal{O}(|\mathcal{V}| log |\mathcal{V}| + |\mathcal{E}|)$. The complexity of Tranformer is $\mathcal{O}(l_{2}|\mathcal{V}|^{2}dh)$. Third, the complexity of cross-propogation is $\mathcal{O}(|\mathcal{V}_{1}||\mathcal{V}_{2}|)$.

\section{Comparison with Graphormer}
Grphormer \cite{ying2021transformers} is a SOTA graph transformer model on the recent OGB-LSC \cite{hu2021ogb} quantum chemistry regression (i.e.,PCQM4M-LSC) challenge, which is currently the biggest graph-level prediction dataset. We also use it in our computation paradigm, and the results are shown in Table. \ref{ggg}. To save computational resources and avoid overfitting, we set the number of layers to 5, the hindden dimension to 128, and the number of heads to 8 in Graphormer. We find that GCwT is more suitable for graph similarity computation tasks since Graformer mainly focuses on the graph-level prediction task.

\makeatletter\def\@captype{table}\makeatother
\begin{center}
\caption{The results of Graphormer. A, L, P denote AIDS, LINUX, PTC\_MM.}
\label{ggg}
\begin{tabular}{l|ccc|cccccc}
\hline
\multicolumn{1}{c|}{\multirow{2}{*}{Datasets}} & \multirow{2}{*}{A} & \multirow{2}{*}{L} & \multirow{2}{*}{P} & \multicolumn{3}{c|}{FFmpeg}                                                                               & \multicolumn{3}{c}{OpenSSL}                                                                              \\
\multicolumn{1}{c|}{}                          &                       &                        &                          & \multicolumn{1}{l}{{[}3, 2H{]}} & \multicolumn{1}{l}{{[}20, 2H{]}} & \multicolumn{1}{l|}{{[}50, 2H{]}} & \multicolumn{1}{l}{{[}3, 2H{]}} & \multicolumn{1}{l}{{[}20, 2H{]}} & \multicolumn{1}{l}{{[}50, 2H{]}} \\ \hline
                                               & \multicolumn{3}{c|}{mse($\times 10^{-2}$)}                                & \multicolumn{6}{c}{AUC score}                                                                                                                                                                                        \\ \hline
Gomer                                 & 1.49                  & 0.31                   & 1.15                     & 97.95                            & 98.41                             & \multicolumn{1}{c|}{98.03}         & 97.24                            & 98.17                             & 97.74                             \\
Ours                                         & 0.30                   & 0.02                   & 0.71                     & 98.49                            & 99.36                             & \multicolumn{1}{c|}{99.49}         & 98.34                            & 99.14                             & 99.26                             \\ \hline
\end{tabular}
\end{center}


\section{The number of parameters}
A comparison of the number of parameters of our method with other baselines can be found in Table \ref{number of parameters}.

\begin{table}[]
\centering
\caption{The scope of hyperparameter search.}
\label{scope_p}
\begin{tabular}{c|c}
Parameter            & Values                                 \\ \hline
\# GCN layer         & 3                                      \\
GCN hidden dimension & \{128, 256, 512\}                      \\
\# Transformer head  & 8                                      \\
\# Transformer layer & \{2, 4, 6, 8\}                         \\
Batch size           & 128 (regression) / 32 (classification) \\
Learning rate        & 0.001                                  \\
\# Epoch             & 100 (regression) / 30 (classification)
\end{tabular}
\end{table}

\begin{table}[]
\centering
\caption{The results of hyperparameter search.}
\label{hyperparameter}
\begin{tabular}{ccccccc}
\hline
                         & \multicolumn{3}{c}{FFmpeg}                   & \multicolumn{3}{c}{OpenSSL}                  \\
                         & {[}3, 200{]} & {[}20, 200{]} & {[}50, 200{]} & {[}3, 200{]} & {[}20, 200{]} & {[}50, 200{]} \\ \hline
\multicolumn{1}{c|}{$l$} & 8            & 6             & 6             & 4            & 8             & 8             \\
\multicolumn{1}{c|}{$d$} & 256          & 128           & 128           & 256          & 256           & 256           \\ \hline
                         & \multicolumn{6}{c}{MCS metric}                                                              \\
                         & AIDS         & LINUX         & PTC\_MM       & BA100        & BA200         & BA300         \\ \hline
\multicolumn{1}{c|}{$l$} & 2            & 2             & 3             & 3            & 2             & 4             \\
\multicolumn{1}{c|}{$d$} & 128          & 256           & 128           & 256          & 128           & 128           \\ \hline
                         & \multicolumn{6}{c}{GED metric}                                                              \\
                         & AIDS         & LINUX         & PTC\_MM       & BA100        & BA200         & BA300         \\ \hline
\multicolumn{1}{c|}{$l$} & 2            & 2             & 3             & 3            & 2             & 3             \\
\multicolumn{1}{c|}{$d$} & 128          & 256           & 128           & 128          & 128           & 128           \\ \hline
\end{tabular}
\end{table}

\begin{table}[] \centering \caption{The number of parameters} \label{number of parameters} \begin{tabular}{cccccc} \hline Model & EMBAVG & GMN & GraphSim & SimGNN & SMPNN \\ Number of Parameters & 26K & 62K & 2M & 17K & 13K \\ \hline Model & H2MN & GOTSim & Base(ours) & INFMCS(ours) & Graphomer \\ Number of Parameters & 291K & 75K & 18K & 306K & 216M \\ \hline \end{tabular} \end{table}

\bibliographystyle{plain}
\bibliography{re}

\end{document}